\documentclass{llncs}

\usepackage{graphicx}
\usepackage{amsmath,algorithm,algpseudocode,epstopdf}
\usepackage[compatibility=false]{caption}
\usepackage{subcaption}

\DeclareMathOperator*{\argmax}{arg\,max}
\DeclareMathOperator*{\argmin}{arg\,min}

\algnewcommand{\Inputs}[1]{%
	\State \textbf{Inputs:}
	\Statex \hspace*{\algorithmicindent}\parbox[t]{.8\linewidth}{\raggedright #1}
}
\algnewcommand{\Outputs}[1]{%
	\State \textbf{Outputs:}
	\Statex \hspace*{\algorithmicindent}\parbox[t]{.8\linewidth}{\raggedright #1}
}

\begin{document}

\title{Robust Ensemble Classifier Combination Based on Noise Removal with One-Class SVM}

\titlerunning{Robust Ensemble Classifier Combination Based on Noise Removal with One-Class SVM}  
%
\author{Ferhat \"{O}zg\"{u}r \c{C}atak}
\authorrunning{Ferhat \"{O}zg\"{u}r \c{C}atak} 
%
\tocauthor{Ferhat Özgür Çatak}
\institute{T\"{U}B\.{I}TAK B\.{I}LGEM, Cyber Security Institute\\
		Kocaeli/Gebze, Turkey,\\
\email{ozgur.catak@tubitak.gov.tr}}

\maketitle              

\begin{abstract}
In machine learning area, as the number of labeled input samples becomes very large, it is very difficult to build a classification model because of input data set is not fit in a memory in training phase of the algorithm, therefore, it is necessary to utilize data partitioning to handle overall data set. Bagging and boosting based data partitioning methods have been broadly used in data mining and pattern recognition area. Both of these methods have shown a great possibility for improving classification model performance. This study is concerned with the analysis of data set partitioning with noise removal and its impact on the performance of multiple classifier models. In this study, we propose noise filtering preprocessing at each data set partition to increment classifier model performance. We applied Gini impurity approach to find the best split percentage of noise filter ratio. The filtered sub data set is then used to train individual ensemble models. 
\end{abstract}
\section{Introduction}\vspace{-10pt}
It's clear that we collect and store larger amounts of data in databases. The need for efficiently and effectively analyzing and utilizing the information contained in the data has been increasing. Just as big data technologies evolved, the quantity and variety of data has also increased, and becoming more focused on storing every type of data. The main purpose of the storing of the data is intended to obtain information from data using a variety of machine learning methods. One of the primary machine learning techniques is classification, which labels the new samples based on a training set whose class labels are provided \cite{anderson1986machine,ramakrishnan2000database}. Classification methods are applied in various areas such as bioinformatics, pattern recognition, text mining, social network analysis, etc.

In Big Data age, traditional classifier algorithms have new challenges to scaling up in order to address the large-scale data set training. Most of existing classification algorithms assume that the data can fit in a memory in training phase of learning. These algorithms cannot be comfortably implemented to data sets that larger than computer memory capacity. Data partitioning strategy is one of the methods that can be applied to the training of high-dimensional data sets that are used for the building of classification model in order to overcome the input data complexity. In order to prevent the building of weak classification model that emerged from the data chunks, the input set needs to be strengthened through various methods. 

In this study, the noise filtering approach is applied to each individual sub data set to clean noisy input data, then, AdaBoost ensemble method is used to strength the classification model at each data partition.  We applied one-class Support Vector Machine (SVM) method to filter noisy instances from each individual data partition and then AdaBoost ensemble based classification method is used to each individual data partition to increase the model accuracy.

The overall contributions of the study are listed as follows:
\begin{enumerate}
	\item Using data partitioning method, the complexity of input matrix, which is quite high for the single memory, is reduced in this manner.
	\item Each individual sub-set of input matrix is reinforced with noise filtering method using one-class SVM and Gini impurity.
	\item Each sub-set of input matrix is used in the training phase of the different ensemble classifier, so that each instances are considered when building a global classification model. 
\end{enumerate}

Gini impurity is used to calculate the uncertainty about source of input data set. This measure is applied to estimate the degree of information diversity provided by cleaned partition of sub data set.

The remainder of this paper is organized as follows: Section \ref{sec:preliminaries} briefly explains the methods that are used in this work. Section \ref{sec:approach} describes the proposed data cleaning and partitioning method. Section \ref{sec:experiments} gives the experimental results. In Section \ref{sec:conclusion}, we give conclusion and future works.
\section{Preliminaries}\label{sec:preliminaries}
The approach presented in this paper uses one-class SVM algorithm to remove noisy instances, AdaBoost to build ensemble classifier models, and data partitioning to train over all data set instances. All elements are introduced here briefly.
\subsection{One-Class SVM}
SVM \cite{vapnik2000nature} method is used to find classification models using the maximum margin separating hyper plane. Sch\"{o}lkopf et al. \cite{scholkopf2001estimating} proposed a training methodology that handles only one class classification called as "one-class" classification.

\begin{figure}[t!]
\begin{center}
\includegraphics[width=8cm]{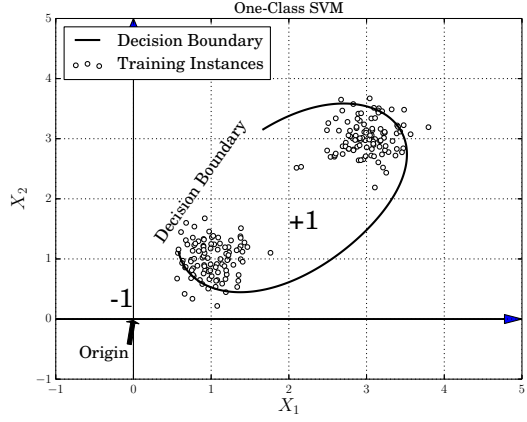}
\caption{One-class SVM. The origin, $(0, 0)$, is the single instance with label $-1$.}
\label{fig:oneclasssvm}
\end{center}
\end{figure}

One-class SVM algorithm is a method used to detect the outliers in the data. Basically the method finds soft boundaries of the data set, and then, model determines whether new instance belongs to this data set or not. Suppose, we are given a data set, $\mathbf{x}_1, \dots \mathbf{x}_m \in X$ drawn from an unknown underlying probability distribution $P$. We are interested in estimating a set $S$ such that the probability that a test point from $P$ lies inside in $S$ with an a priori specified probability value. As shown in Figure \ref{fig:oneclasssvm}, origin is labeled as $-1$, and the all training instances are labeled as $+1$.

Let $S = \{ (\mathbf{x}_i , y_i) | \mathbf{x}_i \in \bbbr^{n}, y \in \{ 1, \dots, K \} \}_{i=1}^m$ be input instances in $\bbbr^{n}$, $\phi : X \rightarrow H$ be a kernel function that maps the input instances to another space. Then standard  SVM method tries to find a hyper plane that solves the separation problem with an optimization problem. The objective function of the SVM classifier is formulated as follows.

\begin{align}
\begin{split}
	min_{\mathbf{w}, \xi, \rho} & \left( \frac{1}{2}||\mathbf{w}||^2 + \frac{1}{mC}\sum_{i} {\xi_i} - \rho \right) \\
	subject \, to &  \\
	& \left( w . \phi (\mathbf{x}_i) \right) \geq \rho - \xi_i  \\
	& \xi_i \geq 0, \forall i = 1,\dots,m
\label{eq:one-class-svm}
\end{split}
\end{align}
where $\mathbf{w}$ is orthogonal to the separating  hyper plane, $C$ is smoothness parameter, $\mathbf{x}_i$ is the $i$-th input instances, $m$ is the total number of input instances, $\xi_i$ are the slack variables, $\rho$ is the distance between origin and separating  hyper plane.

By using Lagrange techniques, $\mathbf{w}$ and $\rho$ are obtained, then the decision function becomes:
\begin{equation}
	f(\mathbf{x}) = sign\left( \left( \mathbf{w} . \phi(\mathbf{x}) \right) - \rho \right)
\end{equation}
\subsection{AdaBoost}
The AdaBoost \cite{freund1995desicion} is a supervised learning algorithm designed to solve classification problems \cite{freund1999short}. The algorithm takes as input a training set $(\mathbf{x}_1, y_1),...,(\mathbf{x}_n, y_n)$ where the input sample $\mathbf{x}_i \in R^p$, and the output value, $y_i$, in a finite space $y\in {1,...K}$. AdaBoost algorithm assumes a set of training data sampled independently and identically distributed (i.i.d.) from some unknown distribution $\mathcal{X}$. 

Given a space of feature vectors $X$ and two possible class labels, $y \in \{-1,+1\}$, AdaBoost goal is to learn a strong classifier $H(\mathbf{x})$ as a weighted ensemble of weak classifiers $h_t(\mathbf{x})$ predicting the label of any instance $\mathbf{x} \in X$ \cite{LandesaVazquez2013101}.
\begin{equation}
	\label{eq:adaboost}
	H(\mathbf{x}) = sign(f(\mathbf{x}))=sign\left(\sum_{t=1}^{T}\alpha_t h_t(\mathbf{x}) \right)
\end{equation}
\subsection{Data Partitioning Strategies}
The use of multiple classifiers, learning methods are applied to base classifiers with different methods. Data partitioning is used a variety of reasons. First reason is the diversity that means uncorrelated base classifiers \cite{kuncheva2005using,dara2010filter}. Another reason is the reducing the input complexity of large-scale data sets \cite{chawla2003distributed}. Last one is to build classifier models for the specific part of the input instances \cite{woods1996combination}.

Data partitioning is basically divided into two different groups; filter based data partitioning and wrapper based data partitioning \cite{blum1997selection}. In wrapper based data partitioning, sub-data sets are created using base classifier outputs \cite{freund1996experiments}. In filter based data partitioning, sub-data sets are created before individual classifiers are trained \cite{breiman1996bagging}.
\section{Proposed Approach}\label{sec:approach}
In this section we provide the details of the proposed noise filter based sub data set training method. The basic idea of noise removing based on one-class SVM technique is introduced in Section \ref{sec:BasicIdea}. The analysis of proposed method is described in Section \ref{sec:analysisofalg}.
\subsection{Basic Idea}\label{sec:BasicIdea}
Our main task is to partition the input data set into sub-data sets, $(X_m,Y_m)$, and, create local classifier ensembles for each sub data chunk. Noise removing process is applied to each individual sub-data set as pre-processing. Weighted voting method is used to combine the each ensemble classifier, and then, a single classifier model is created. Overall of the proposed method is shown in Figure \ref{fig:overall}.



\begin{figure}[h]
	\label{fig:overall}
	\begin{center}
		\includegraphics[width=10cm]{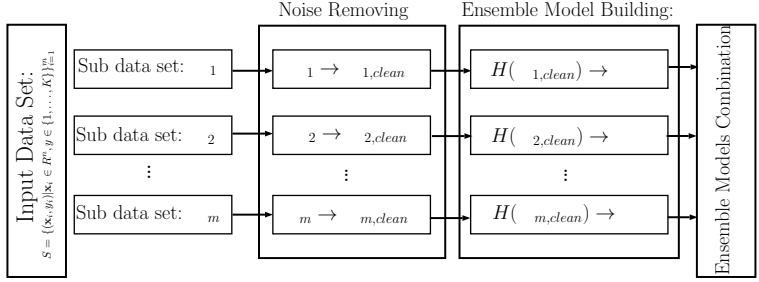}
		\caption{Overall of proposed approach.}
		\label{fig:overall}
	\end{center}
\end{figure}
\subsection{Analysis of the proposed algorithm}\label{sec:analysisofalg}
Kragh et al. showed that ensemble methods of neural networks gets better accuracy performance over unseen examples \cite{Krogh95neuralnetwork}. The main motivation of  this work is the idea that small size classifier ensembles can obtain more accurate classifier model that are comparable to individual classifiers.

In the proposed model, at every sub-data set, there is a set of classifier functions (ensemble classifier), $H^{(m)}$, that acts as a single classification model. The single model at every sub-data set, $m$, is defined as follows:
\begin{equation}
\label{eq:setELM}
H^{(m)}(\mathbf{x}) = \argmax_k \sum_{t=1}^{M}{\alpha_t h_t(\mathbf{x})}
\end{equation}
The selected ensemble classifier models from last phase of our algorithm are combined into one single classification model, $\hat{H}(\mathbf{x})$, using accuracy based majority voting method.
\begin{equation}
\label{eq:setELMFinal}
\hat{H}(\mathbf{x}) = \argmax_k \sum_{i=1}^{m}{\beta H^{(m)}( \mathbf{x})}
\end{equation}
where $\beta$ is the accuracy of ensemble classifier.
\section{Experiments}\label{sec:experiments}
In this section, we perform experiments on real-world data sets from the public available data set repositories. Public data sets are used to evaluate the proposed learning method. Classification models of each data set are compared for accuracy results  without removing noisy samples from them. 
\subsection{Experimental setup}\label{sec:expsetup}
In this section, our approach is applied to five different data sets to verify its model effectivity and efficiency. The data sets are summarized in Table \ref{tbl:dslist}, including cod-rna, ijcnn1, letter, shuttle and SensIT Vehicle. We choose $50$ as the data split size, $m$, and 3 different classification methods including Extra Trees \cite{geurts2006extremely}, k-nn and SVM.
\begin{table}[h]
	\caption{Description of the testing data sets used in the experiments.}
	\label{tbl:dslist}
	\begin{center}
		\begin{tabular}{|c||r|r|c|c|}
			\hline Data set & \#Train & \#Test & \#Classes & \#Attributes \\ 
			\hline \hline cod-rna & 59,535 & 157,413 & 2 & 8 \\ 
			\hline ijcnn1 & 49,990 & 91,701 & 2 & 22 \\ 
			\hline letter & 15,000 & 5,000  & 26 & 16 \\ 
			\hline shuttle & 43,500 & 14,500 & 7 & 9 \\
			\hline SensIT Vehicle & 78,823 & 19,705 & 3 & 100 \\
			\hline 
		\end{tabular} 
	\end{center}
\end{table}
\subsection{Effect of Noise Removing on Input Matrix}
In this section, we show the impact of noise removal pre-processing on the sample data sets. In order to show the noise removing affects, we used the "Gini Impurity" to measure the quality of procedure. Gini approaches deal appropriately with data diversity of a data. The Gini measures the class distribution of variable $\mathbf{y} = \{ y_1, \cdots, y_m \}$. The Gini impurity can be written as :
\begin{equation}
	\label{eqn:gini}
	g = 1 - \sum_{k}{p^2_j}
\end{equation}
where $p_j$ is the probability of class $k$, in data set $\mathcal{D}$. 



\begin{figure}[h!]
	\begin{minipage}[b]{0.5\textwidth}
		\includegraphics[width=1\linewidth]{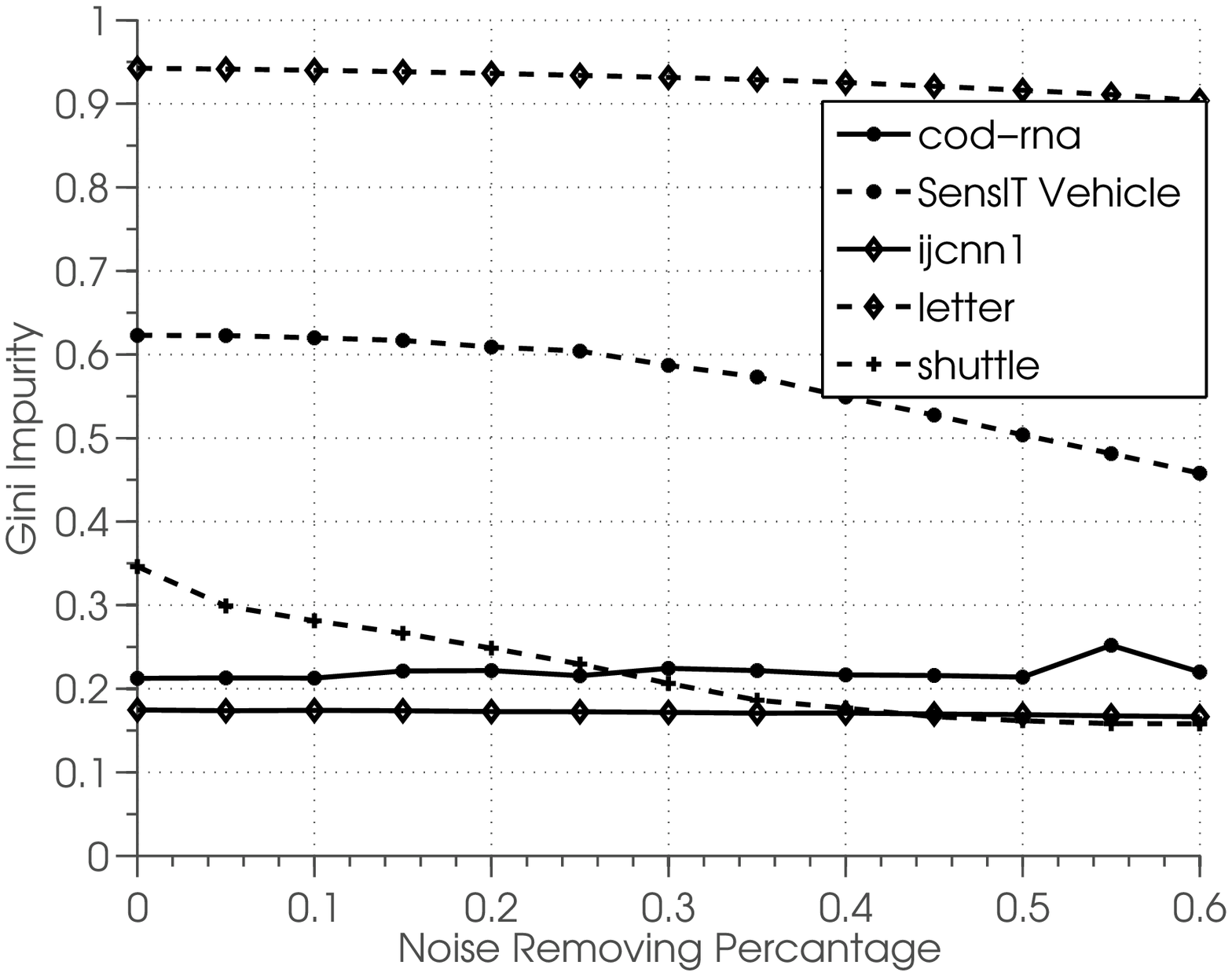}
		\subcaption{Cleaned part.}
		\label{fig:cleanimp}
	\end{minipage}%
	\begin{minipage}[b]{0.5\textwidth}
		\includegraphics[width=1\linewidth]{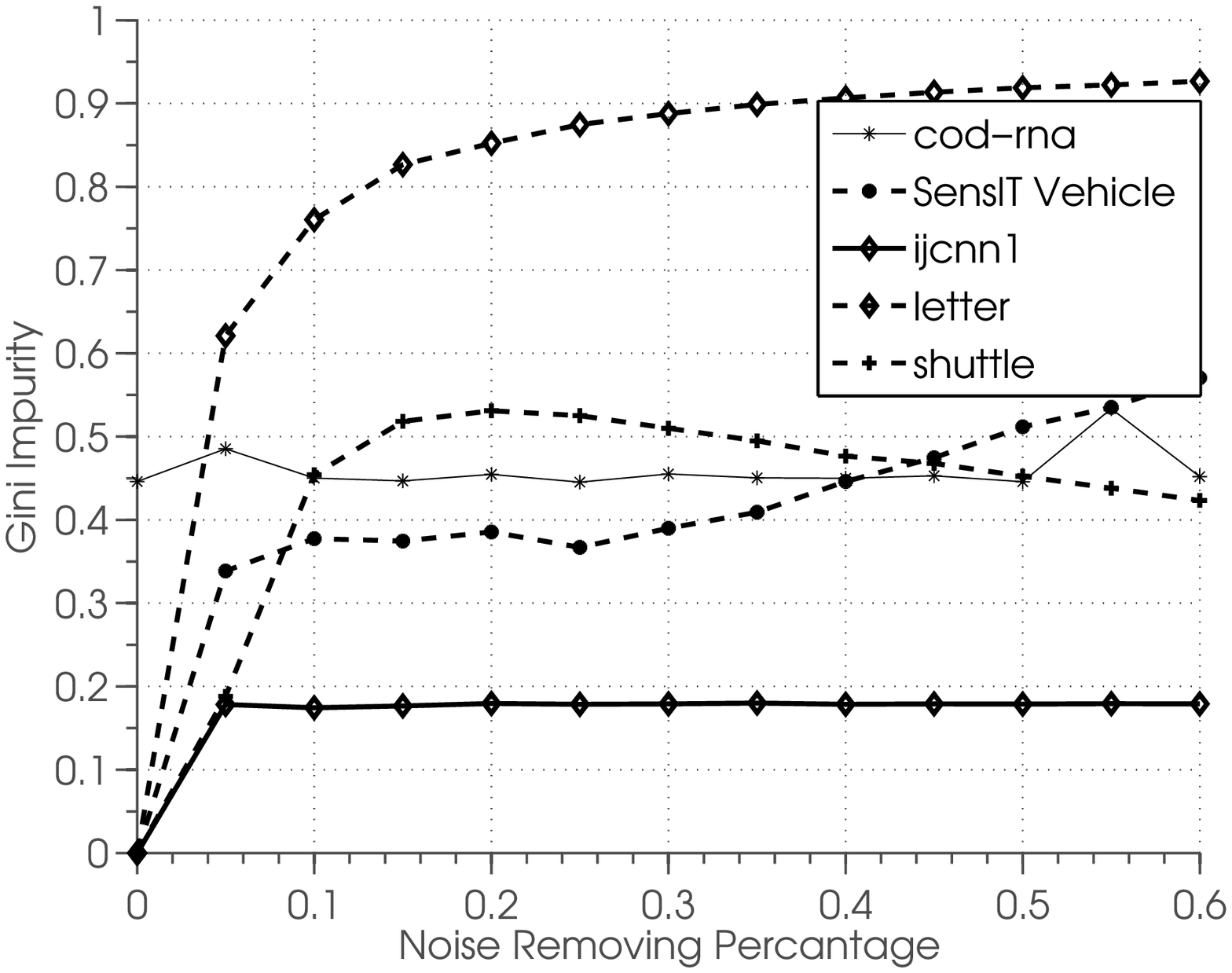}
		\subcaption{Removed part.}
		\label{fig:novelty-impurity}
	\end{minipage}%
\caption{The impact of one-class SVM on the performance on selected data sets in terms of Gini impurity.}
\end{figure}
The cleaning results are shown in Figure \ref{fig:cleanimp} and Figure \ref{fig:novelty-impurity}.  As expected, the Gini impurity value decreases with the cleaning of the noisy instances from the data, and increases on separated noisy data. Our aim is to minimizing the Gini impurity on clean data set, $\mathbf{X}_{clean}$, and maximizing the value on noisy data set $\mathbf{X}_{noisy}$. As a result, this division ratio which minimizes the ratio between the two values was regarded as the optimum value.
\begin{equation}
	\label{eqn:splitsize}
	Split \, Percantage = \argmin_p \frac{Gini(\mathbf{X}_{clean},p)}{ Gini(\mathbf{X}_{noisy},p)} 
\end{equation}
Table \ref{tbl:bestper} shows the best Gini impurity performances of each data set used in our experiments.\vspace{-30pt}
\begin{table}[h!]
	\caption{The best noise removal percentages of each data sets.}
	\label{tbl:bestper}
	\begin{center}
		\begin{tabular}{r@{\quad}rrrr}
			\hline
			\multicolumn{1}{l}{\rule{0pt}{12pt}
				Data Sets}& Percentage & $Gini(\mathbf{X}_{clean})$ & $Gini (\mathbf{X}_{noisy})$ & $\frac{Gini(\mathbf{X}_{clean})}{ Gini(\mathbf{X}_{noisy})} $  \\[2pt]
			\hline\rule{0pt}{12pt}
			cod-rna & 0,55 & 0,331344656 & 0,56831775 & 0,583027112 \\
			SensIT Vehicle & 0,60 & 0,461155751 & 0,568847982 & 0,810683638 \\
			ijcnn1 & 0,60 & 0,167652825 & 0,179397223 & 0,934534117 \\
			letter & 0,60 & 0,9039108 & 0,926430562 & 0,975691905 \\
			shuttle & 0,30 & 0,214142833 & 0,506938229 & 0,422423918  \\[2pt]
			\hline
		\end{tabular}
	\end{center}
\end{table}\vspace{-40pt}
\subsection{Simulation Results}\label{sec:simresults}
The process of the experiments are as follows: Firstly, we trained our data sets without using noise removal. Then we perform classification on test data sets, and calculate the accuracy of classifiers. We repeated the experiments 50 times, and average accuracy is calculated. Table \ref{tbl:dsresults} shows the average accuracy of each example data sets with and without noise removing using one-class SVM method.

As can be seen on Table \ref{tbl:dsresults}, the noise removing based partitioned proposed algorithm significantly outperforms the splitted classifier building in most cases.
\vspace{-20pt}
\begin{table}[h]
	\caption{Classification performance on example datasets using One-Class SVM noise removing and without removing for the proposed learning algorithm.}
	\label{tbl:dsresults}
	\begin{center}
		\begin{tabular}{|c||r|r||r|r||r|r|}
			\hline  & \multicolumn{2}{|c|}{\textbf{Extra trees}}  & \multicolumn{2}{|c|}{\textbf{K-nn}}  & \multicolumn{2}{|c|}{\textbf{SVM}} \\ 
			\hline \textbf{Data set} & All & Clean & All & Clean & All & Clean \\
			\hline \hline cod-rna & 0,75929 & 0,78652 & 0,91955 & 0,93513 & 0,88553 & 0,89806 \\
			\hline ijcnn1 & 0,69758 & 0,72175 & 0,75533 & 0,77833 & 0,82989 & 0,84326 \\
			\hline letter & 0,91255 & 0,90853 & 0,90594 & 0,90318 & 0,92121 & 0,9199 \\
			\hline shuttle & 0,47801 & 0,44792 & 0,20262 & 0,26974 & 0,62301 & 0,63939 \\
			\hline SensIT Vehicle & 0,9602 & 0,90558 & 0,87904 & 0,88266 & 0,99232 & 0,99174 \\
			\hline
	\end{tabular} 
	\end{center}
\end{table}\vspace{-20pt}
\section{Conclusions}\label{sec:conclusion}\vspace{-10pt}
In this paper, we have introduced a novel data partitioning based classifier building method, which improves the sub data sets with removing the noisy instances using one-class SVM and find best noise removing ratio with Gini impurity value.  We carried out a series of computer experiments to find a global ensemble classifier and the performance of our proposed method. The training process of a partitioned data set is simple, fast and final classifier model handle overall training instances. Our experimental results show that the memory requirement of training phase reduced remarkably, and the accuracy increased by using the noise removal process. The proposed method is a practical multiple ensemble classifier training model to classify large-scale data sets.

In the future work, our plan is to study different noise removing methods to clean sub data set. We plan adaptive noise removing ratio to make our method as autonomous as possible.\vspace{-10pt}
%
%
%
%
%
%
%
%

\bibliographystyle{splncs}
\bibliography{references}

\end{document}